\def\year{2021}\relax
\documentclass[letterpaper]{article} 
\usepackage{aaai21}  
\usepackage{times}  
\usepackage{helvet} 
\usepackage{courier}  
\usepackage[hyphens]{url}  
\usepackage{graphicx} 
\usepackage{natbib}  
\usepackage{caption} 
\usepackage[switch]{lineno} 
\usepackage[ruled,vlined]{algorithm2e}
\usepackage{amsmath}
\usepackage{array}
\usepackage{amssymb}
\usepackage{amsfonts}
\usepackage{booktabs}
\usepackage{multirow}
\usepackage{multicol}
\usepackage{mathrsfs}
\usepackage{amsthm}
\usepackage{xcolor}
\newtheorem{theorem}{Theorem}
\newcommand{\uwave}{\textcolor{red}}
\newcommand{\newcite}[1]{\citeauthor{#1}~(\citeyear{#1})}
\newcommand{\bfx}{\textbf{x}}
\newcommand{\bfz}{\textbf{z}}

\begin{document}
%
\title{Learning from the Best: Rationalizing Prediction by\\ Adversarial Information Calibration}
\author{
    Lei Sha,\textsuperscript{1}  \ \ Oana-Maria  Camburu,\textsuperscript{1,2}  \ \ Thomas Lukasiewicz\textsuperscript{1,2}
    \\
}
\affiliations{
    \textsuperscript{1}Department of Computer Science, University of Oxford, UK \qquad
  \textsuperscript{2}Alan Turing Institute, London, UK \\
  \texttt{firstname.lastname@cs.ox.ac.uk}

}

\maketitle
\begin{abstract}
\begin{quote}
Explaining the predictions of AI models is paramount in safety-critical applications, such as in legal or medical domains.
One form of explanation for a prediction is an extractive rationale, i.e., a subset of features of an instance that lead the model to give its prediction on the instance. 
Previous works on generating extractive rationales usually employ a two-phase model: a selector that selects the most important features (i.e., the rationale) followed by a predictor that makes the prediction based exclusively on the selected features.
One disadvantage of these works is that the main signal for learning to select features comes from the comparison of the answers given by the predictor and the ground-truth answers. In this work, we propose to squeeze more information from the predictor via an information calibration method. More precisely, we train two models jointly: one is a typical neural model that solves the task at hand in an accurate but black-box manner, and the other is a selector-predictor model that additionally produces a rationale for its prediction. The first model is used as a guide to the second model. We use an adversarial-based technique to calibrate the information extracted by the two models such that the difference between them is an indicator of the missed or over-selected features. In addition, for natural language tasks, we propose to use a language-model-based regularizer to encourage the extraction of fluent rationales.
Experimental results on a sentiment analysis task as well as on three tasks from the legal domain show the effectiveness of our approach to rationale extraction.

\end{quote}
\end{abstract}

\section{Introduction}
Although deep neural networks have recently been contributing to state-of-the-art advances in various areas \cite{Krizhevsky:2012:ICD:2999134.2999257, hinton2012deep, DBLP:journals/corr/SutskeverVL14}, 
such black-box models may not be deemed reliable in situations where safety needs to be guaranteed, such as legal judgment prediction and medical diagnosis. Interpretable deep neural networks are a promising way to increase the reliability of neural models~\cite{sabour2017dynamic}. To this end, extractive rationales, i.e., subsets of features of instances on which models rely for their predictions on the instances, can be used as evidence for humans to decide whether or not to trust a predicted result and, more generally, to trust a~model. 

Previous works mainly use selector-predictor types of neural models to provide extractive rationales, i.e., models composed of two modules: (i) a \textit{selector} that selects a subset of important features, and (ii) a  \textit{predictor} that makes a prediction based solely on the selected features. For example, \newcite{yoon2018invase} and \newcite{lei2016rationalizing} use a selector network to calculate a selection probability for each token in a sequence, then sample a set of tokens that is exclusively passed to the predictor. 

An additional typical desideratum in natural language pro\-cessing (NLP) tasks is that the selected tokens form a semantically fluent rationale. To achieve this, \newcite{lei2016rationalizing} added a non-differential regularizer that encourages any two adjacent tokens to be simultaneously selected or unselected. 
\newcite{bastings2019interpretable} further improved the quality of the rationales by using a Hard Kuma regularizer that also encourages any two adjacent tokens to be selected or unselected together. 

\begin{figure}[!t]
\begin{center}
\includegraphics[width=\linewidth]{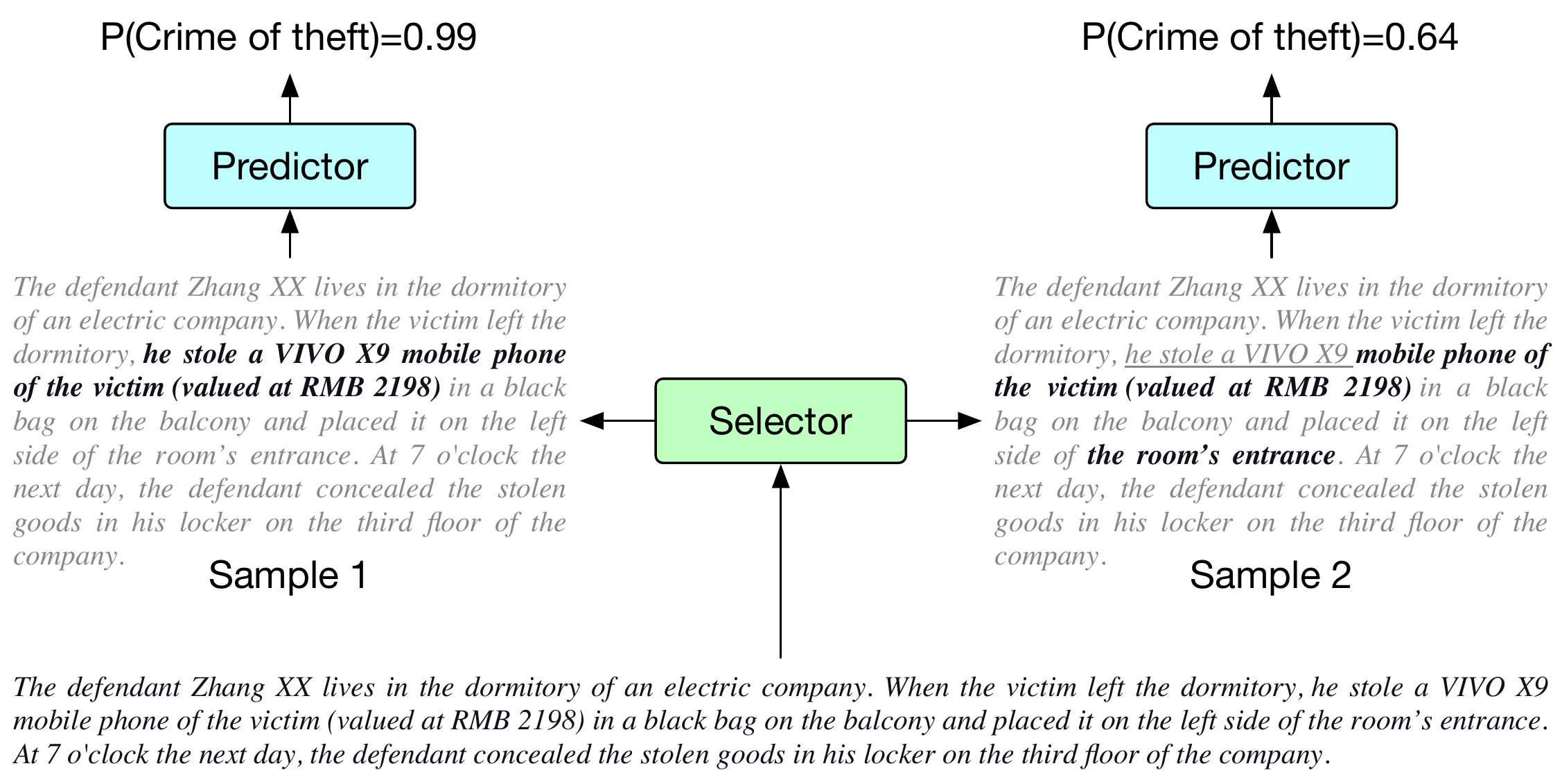}\\[0.75ex]
\caption{A sample rationale in legal judgement prediction. The human-provided rationale is shown in bold in Sample~1. In Sample 2, the selector missed the key information ``he stole a VIVO X9'', but the predictor only tells the selector that the whole extracted rationale (in bold) is not so informative, by producing a low probability of the correct crime. }
\label{fig:intro}
\end{center}
\end{figure}

One drawback of previous works is that the learning signal for both the selector and the predictor comes mainly from comparing the prediction of the selector-predictor model with the ground-truth answer.
Therefore, the exploration space to get to the correct rationale is large, decreasing the chances of converging to the optimal rationales and predictions. 
Moreover, in NLP applications, the regularizers commonly used for achieving fluency of rationales treat all adjacent token pairs in the same way. This often leads to the selection of unnecessary tokens due to their adjacency to informative~ones. 

In this work, we first propose an alternative method to rationalize the predictions of a neural model. Our method aims to squeeze more information from the predictor in order to guide the selector in selecting the rationales. Our method trains two models: a ``guider" model that solves the task at hand in an accurate but black-box manner, and a selector-predictor model that solves the task while also providing rationales. We use an adversarial-based method to encourage the final information vectors generated by the two models to encode the same information. We use an information bottleneck technique in two places: (i)~to encourage the features selected by the selector to be the least-but-enough features, and (ii)~to encourage the final information vector of the guider model to also contain the least-but-enough information for the prediction.   
Secondly, we propose using language models as regularizers for rationales in natural language understanding tasks. A language model (LM) regularizer encourages rationales to be fluent subphrases, which means that the rationales are formed by consecutive tokens while avoiding unnecessary tokens to be selected simply due to their adjacency to informative tokens. 
The effectiveness of our LM-based regularizer is proved by both mathematical derivation and experiments. All the further details are given in the Appendix of the extended (ArXiv) paper.

Our contributions are briefly summarized as follows:
\begin{itemize}
\item We introduce an adversarial approach to rationale extraction for neural predictions, which calibrates the information between a guider and a selector-predictor model, such that the selector-predictor model learns to mimic a typical neural model while additionally providing rationales.


\item We propose a language-model-based regularizer to encourage the sampled tokens to form fluent rationales. 

\item We experimentally evaluate our method on a sentiment analysis dataset with ground-truth rationale annotations, and on three tasks of a legal judgement prediction dataset, for which we conducted human evaluations of the extracted rationales. The results show that our method improves over the previous state-of-the-art models in precision and recall of rationale extraction without sacrificing the prediction performance.
\end{itemize}

\section{Approach}
Our approach is composed of a selector-predictor architecture, in which  we use an information bottleneck technique to restrict the number of selected features, and a guider model, for which we again use the information bottleneck technique to restrict the information in the final feature vector. Then, we use an adversarial method to make the guider model guide the selector to select least-but-enough features. Finally, we use an LM regularizer to make the selected rationale semantically fluent.

\subsection{InfoCal: Selector-Predictor-Guider with Information Bottleneck}

The high-level architecture of our model, called InfoCal, is shown in Fig.~\ref{fig:arch}. Below, we detail each of its components.  

\subsubsection{Selector.} For a given instance $(\bfx,y)$, $\bfx$ is the input with $n$ features $\bfx = (x_1, x_2, \ldots, x_n)$, and $y$ is the ground-truth corresponding label. The selector network $\text{Sel}(\tilde{\bfz}_\text{sym}|\bfx)$ takes $\bfx$ as input and outputs $p(\tilde{\bfz}_\text{sym}|\bfx)$, which is a sequence of probabilities $(p_i)_{i=1,\ldots,n}$ representing the probability of choosing each feature $x_i$ as part of the rationale. 

Given the sampling probabilities, a subset of features is sampled using the Gumbel softmax~\cite{jang2016categorical}, which provides a differentiable sampling process: 
\begin{align}
u_i&\sim U(0,1),\quad g_i=-\log(-\log(u_i))\\
m_i&=\frac{\exp((\log(p_i)+g_i)/\tau)}{\sum_j\exp((\log(p_j)+g_j)/\tau)},\label{eq:maski}
\end{align}
where $U(0,1)$ represents the uniform distribution between $0$ and $1$, and $\tau$ is a temperature hyperparameter. 
Hence, we obtain the sampled mask $m_i$ for each feature $x_i$, and the vector symbolizing the rationale $\tilde{\bfz}_\text{sym}=(m_1x_1,\ldots,m_nx_n)$. Thus, $\tilde{\bfz}_\text{sym}$ is the sequence of discrete selected symbolic features forming the rationale.

\subsubsection{Predictor.} The predictor takes as input the rationale $\tilde{\bfz}_\text{sym}$ given by the selector, and outputs the prediction $\hat{y}_{sp}$. 
In the selector-predictor part of InfoCal, the input to the predictor is the multiplication of each feature $x_i$ with the sampled mask $m_i$. The predictor first calculates a dense feature vector $\tilde{\bfz}_\text{nero}$,\footnote{Here, ``nero'' stands for neural feature (i.e., a neural vector representation) as opposed to a symbolic input feature.} then uses one feed-forward layer and a softmax layer to calculate the probability distribution over the possible predictions: 
\begin{align}
\tilde{\bfz}_\text{nero}&=\text{Pred}(\tilde{\bfz}_\text{sym})\\
p(\hat{y}_{sp}|\tilde{\bfz}_\text{sym}) &= \text{Softmax}(W_p\tilde{\bfz}_\text{nero}+b_p).\label{eq:pyn}
\end{align}
As the input is masked by $m_i$, the prediction $\hat{y}_{sp}$ is made~exclusively based on the features selected by the selector. The loss of the selector-predictor model is the cross-entropy loss:
\begin{equation}\label{eq:lsp}
\begin{small}
\begin{aligned}
L_{sp}&=-\frac{1}{K}\sum_k\log p( y^{(k)}_\text{sp}|\bfx^{(k)})\\
&=-\frac{1}{K}\sum_k\log \mathbb E_{\text{Sel}(\tilde{\bfz}_\text{sym}^{(k)}|\bfx^{(k)})}p(y^{(k)}_\text{sp}|\tilde{\bfz}_\text{sym}^{(k)})\\
&\le -\frac{1}{K}\sum_k\mathbb E_{\text{Sel}(\tilde{\bfz}_\text{sym}^{(k)}|\bfx^{(k)})}\log p(y^{(k)}_\text{sp}|\tilde{\bfz}_\text{sym}^{(k)}), 
\end{aligned}
\end{small}
\end{equation}
where $K$ represents the size of the training set, the superscript (k) denotes the k-th instance in the training set, and the inequality follows from Jensen's inequality.

\subsubsection{Guider.} 
\begin{figure}[!t]
\begin{center}
\includegraphics[width=\linewidth]{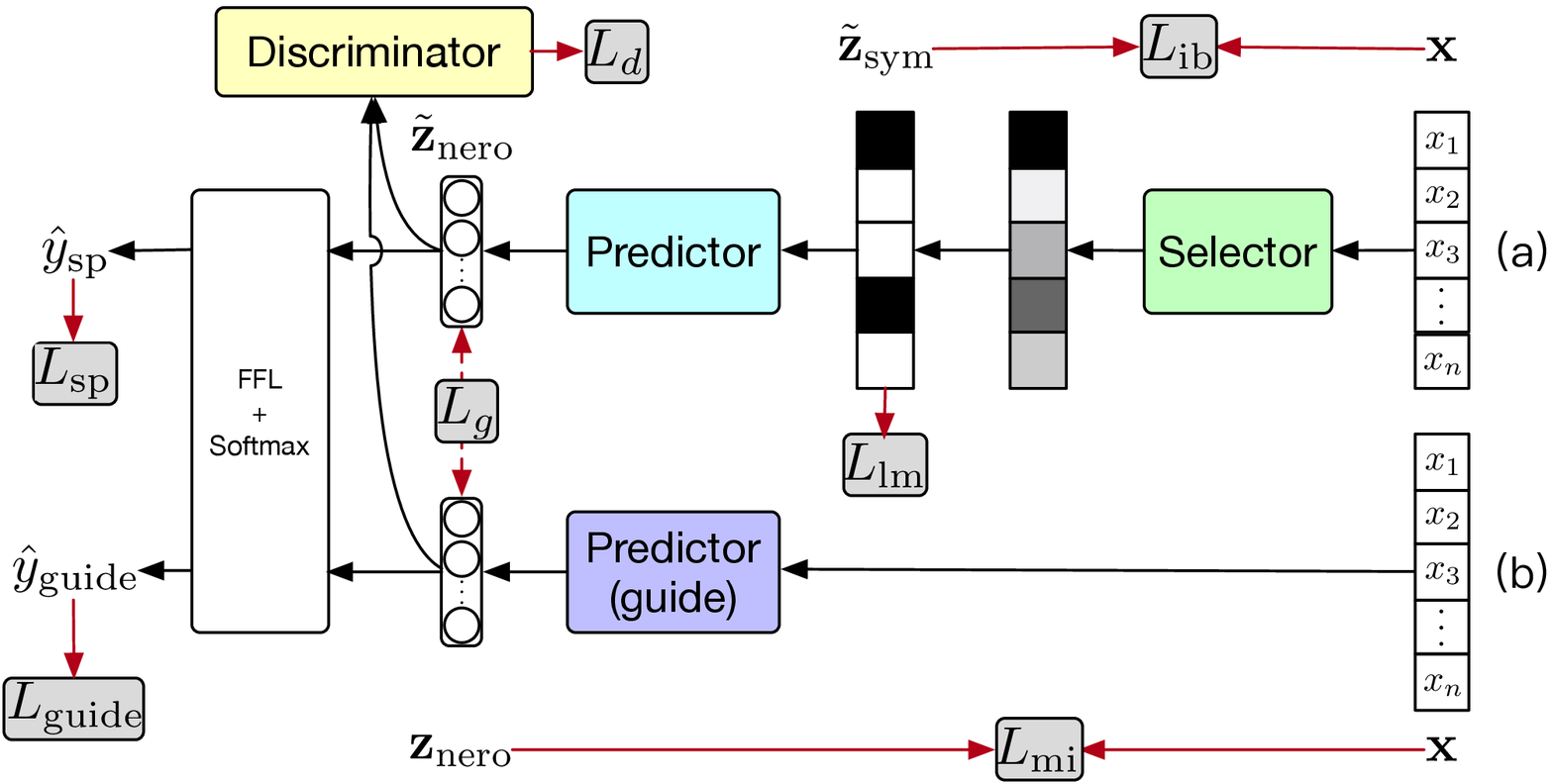}\\
\caption{Architecture of InfoCal: the grey round boxes stand for the losses, and the red arrows indicate the data required for the calculation of losses. FFL is an abbreviation for feed-forward layer.}
\label{fig:arch}
\end{center}
\end{figure}

To guide the rationale selection of the selector-predictor model, we train a \textit{guider} model, denoted Pred$_G$, which receives the full original input $\bfx$ and transforms it into a dense feature vector $\bfz_\text{nero}$, using the same predictor architecture but different weights, as shown in Fig.~\ref{fig:arch}. We generate the dense feature vector in a variational way, which means that we first generate a Gaussian distribution according to the input $\bfx$, from which we sample a vector $\bfz_\text{nero}$: 
\begin{align}
h&=\text{Pred}_G(\bfx),\quad\mu=W_mh+b_m,\quad\sigma=W_sh+b_s \\
u&\sim \mathcal N(0,1), \quad \bfz_\text{nero} = u\sigma + \mu\\
p&(\hat y_\text{guide}|\bfz_\text{nero}) = \text{Softmax}(W_p\bfz_\text{nero}+b_p).
\end{align}
We use the reparameterization trick of Gaussian distributions to make the sampling process differentiable~\cite{kingma2013auto}. Note that we share the parameters $W_p$ and $b_p$ with those in Eq.~\ref{eq:pyn}.

The guider model's loss $L_\text{guide}$ is as follows:
\begin{equation}\label{eq:full}
\begin{aligned}
L_\text{guide}&=-\frac{1}{K}\sum_k\log p(y^{(k)}_\text{guide}|\bfx^{(k)})\\
&\le-\frac{1}{K}\sum_k\mathbb E_{p(\bfz_\text{nero}|\bfx^{(k)})}\log p(y^{(k)}_\text{guide}|\bfz_\text{nero}^{(k)}), 
\end{aligned}
\end{equation}
where the inequality again follows from Jensen's inequality. The guider and the selector-predictor are trained jointly.
\subsubsection{Information Bottleneck.} 
To guide the model to select the least-but-enough information, we employ an information bottleneck technique \cite{li-eisner-2019-specializing}. We aim to minimize $I(\bfx, \tilde{\bfz}_\text{sym}) - I(\tilde{\bfz}_\text{sym},y)$\footnote{$I(a,b) = \int_a\int_bp(a,b)\log\frac{p(a,b)}{p(a)p(b)} \,{=}\, \mathbb E_{a,b}[\frac{p(a|b)}{p(a)}]$ denotes  the mutual information between the variables $a$ and  $b$.}, where the former term encourages the selection of few features, and the latter term encourages the selection of the necessary features.  As $I(\tilde{\bfz}_\text{sym},y)$ is implemented by $L_{sp}$ (the proof is given in Appendix \ref{the:0} in the extended paper), we only need to minimize the mutual information $I(\bfx, \tilde{\bfz}_\text{sym})$:
\begin{align}\label{eq:Isym}
I(\bfx, \tilde{\bfz}_\text{sym})=\mathbb E_{\bfx, \tilde{\bfz}_\text{sym}}\Big[\log\frac{p(\tilde{\bfz}_\text{sym}|\bfx)}{p(\tilde{\bfz}_\text{sym})}\Big].
\end{align}

However, there is a time-consuming term $p(\tilde{\bfz}_\text{sym})=\sum_{\bfx}p(\tilde{\bfz}_\text{sym}|\bfx)p(\bfx)$, which needs to be calculated by a loop over all the instances $\bfx$ in the training set. Inspired by \newcite{li-eisner-2019-specializing}, we replace this term with a variational distribution $r_\phi(z)$ and obtain an upper bound of Eq.~\ref{eq:Isym}: $I(\bfx, \tilde{\bfz}_\text{sym}) \le \mathbb E_{\bfx, \tilde{\bfz}_\text{sym}}\Big[\log\frac{p(\tilde{\bfz}_\text{sym}|\bfx)}{r_\phi(z)}\Big]$. Since $\tilde{\bfz}_\text{sym}$ is a sequence of binary-selected features, we sum up the mutual information term of each element of $\tilde{\bfz}_\text{sym}$ as the information bottleneck loss:
\begin{align}
L_\text{ib}=\sum_i\sum_{\tilde{z}_i} p(\tilde{z}_i|\bfx)\log\frac{p(\tilde{z}_i|\bfx)}{r_\phi(z_i)}, 
\end{align}
where $\tilde{z}_i$ represents whether to select the $i$-th feature: $1$ for selected, $0$ for not selected.

To encourage $\bfz_\text{nero}$ to contain the least-but-enough information in the guider model, we again use the information bottleneck technique. 
Here, we minimize $I(\bfx, \bfz_\text{nero}) - I(\bfz_\text{nero},y)$. Again, $I(\bfz_\text{nero},y)$ can be implemented by $L_\text{guide}$.
Due to the fact that $\bfz_\text{nero}$ is sampled from a Gaussian distribution, the mutual information has a closed-form upper bound:
\begin{equation}\label{eq:mi}
\begin{aligned}
L_\text{mi}&=I(\bfx, \bfz_\text{nero})\le \mathbb E_{\bfz_\text{nero}}\Big[\log\frac{p(\bfz_\text{nero}|\bfx)}{p(\bfz_\text{nero})}\Big] =\\
&=0.5(\mu^2+\sigma^2-1-2\log\sigma).
\end{aligned}
\end{equation}
The derivation is in Appendix~\ref{proof:mi}  in the extended paper.

\subsection{Calibrating Key Features via Adversarial Training}
We would like to tell the selector what kind of information is still missing or has been wrongly selected. Since we already use the information bottleneck principal to encourage $\bfz_\text{nero}$ to encode the information from the least-but-enough features, if we also require $\tilde{\bfz}_\text{nero}$ and $\bfz_\text{nero}$ to encode the same information, then we would encourage the selector to select the least-but-enough discrete features. 
To achieve this, we use an adversarial-based training method.
Thus, we employ an additional discriminator neural module, called~$D$, which takes as input either $\tilde{\bfz}_\text{nero}$ or $\bfz_\text{nero}$ and outputs 0 or 1, respectively. The discriminator can be any differentiable neural network. The generator in our model is formed by the selector-predictor that outputs $\tilde{\bfz}_\text{nero}$. 
The losses associated with the generator and discriminator are:
\begin{align}
L_d&=-\log D(\bfz_\text{nero}) + \log D(\tilde{\bfz}_\text{nero})\label{eq:D}\\
L_g&=- \log D(\tilde{\bfz}_\text{nero}).
\end{align}


\subsection{Regularizing Rationales with Language Models}


For NLP tasks, it is often desirable that a rationale is formed of 
fluent subphrases
\cite{lei2016rationalizing}. To this end, previous works propose regularizers that bind the adjacent tokens to make them be simultaneously sampled or not. For example, \newcite{lei2016rationalizing} proposed a non-differentiable regularizer trained using REINFORCE~\cite{williams1992simple}. To make the method differentiable, \newcite{bastings2019interpretable} applied the Kuma-distribution to the regularizer.
However, they treat all pairs of adjacent tokens in the same way, although some adjacent tokens have more priority to be bound than others, such as  ``He stole'' or ``the victim'' rather than ``. He'' or ``) in'' in Fig.~\ref{fig:intro}. 


We propose a novel differentiable regularizer for extractive rationales that is based on a pre-trained language model, thus encouraging both consecutiveness and fluency of tokens in the extracted rationale. 
The LM-based regularizer is implemented as follows:
\begin{equation}\label{eq:lm}
    L_\text{lm} = -\sum_im_{i-1}\log p_{lm}(m_ix_i|\bfx_{<i}),
\end{equation}
where the $m_i$'s are the masks obtained in Eq.~\ref{eq:maski}. 
Note that non-selected tokens are masked instead of deleted in this regularizer. The language model 
can have any architecture. 

First, we note that $L_\text{lm}$ is differentiable. Secondly, the following theorem guarantees that $L_\text{lm}$ encourages consecutiveness of selected tokens.

\begin{theorem}\label{the:1}
If the following is satisfied for all $i,j$: 
\begin{itemize}
    \item $m'_i<\epsilon \ll 1-\epsilon< m_i$, \,$0<\epsilon<1$, and 
    \item $\big|p(m'_ix_i|x_{<i})-p(m'_jx_j|x_{<j})\big|<\epsilon$,
\end{itemize}
then the following two inequalities hold:\\
 (1) $L_\text{lm}(\ldots,m_k,\ldots, m'_{n})<L_\text{lm}(\ldots,m'_k,\ldots, m_{n})$.\\
 (2) $L_\text{lm}(m_1, \ldots,m'_k,\ldots)>L_\text{lm}(m'_1,\ldots,m_k,\ldots)$.
\end{theorem}
The theorem says that for the same number of selected tokens, if they are consecutive, then they will get a lower $L_\text{lm}$ value.
Its proof is given in Appendix \ref{proof:1}  in the extended paper. 
The pre-training procedure of the language model is shown in  Appendix \ref{lmvec}  in the extended paper.

\begin{table*}[!t]
\centering
\resizebox{0.9\linewidth}{!}{
\begin{tabular}{|l|c|c|c|c|c|c|c|c|c|c|c|c|}
\hline
\multirow{2}{*}{Method} & \multicolumn{4}{c|}{Appearance} & \multicolumn{4}{c|}{Smell}& \multicolumn{4}{c|}{Palate}\\\cline{2-13}
           & P& R& F & \% selected      &  P& R& F  & \% selected      &  P& R& F  & \% selected \\\hline
   Attention  &80.6   &35.6   & 49.4  &13  &88.4  &20.6 &33.4   &7   &65.3  &35.8  &46.2   &7 \\\hline
   Bernoulli  &96.3   &56.5   &71.2    &14  &95.1  &38.2 &54.5   &7   &80.2  &53.6  &64.3  &7\\\hline
   HardKuma  &98.1   &65.1    &78.3    &13  &\textbf{96.8}  &31.5 &47.5   &7   &\textbf{89.8}  &48.6  &63.1  &7\\\hline\hline
   InfoCal  &\textbf{98.5}   &\textbf{73.2} &\textbf{84.0}  &13  &95.6  &\textbf{45.6} &\textbf{61.7}   &7   &89.6  &\textbf{59.8}  &\textbf{71.7}  &7\\\hline
   InfoCal(HK)  &97.9   &71.7 &  82.8   &13  &94.8  &42.3 & 58.5       &7   &89.4  &56.9  & 69.5     &7\\\hline
   InfoCal$-L_\text{adv}$  &97.3      &67.8   & 79.9 &13 &94.3  &34.5 &50.5   &7   &89.6  &51.2  &65.2  &7\\\hline
   InfoCal$-L_\text{lm}$  &79.8   &54.9    &65.0  &13  &87.1  &32.3 &47.1   &7  &83.1  &47.4  &60.4  &7\\\hline
\end{tabular}
}
\smallskip 
 \caption{Precision, recall, and F1-score of selected rationales for the three aspects of BeerAdvocate. In bold, the best performance. ``\% selected'' means the average percentage of tokens selected out of the total number of tokens per instance.}
\label{tab:beer_rational}
\end{table*}%

\begin{table*}[!t]
\centering
\resizebox{0.9\linewidth}{!}{
\begin{tabular}{|c|m{25cm}|}
\hline
Gold & \textcolor{red}{clear , burnished copper-brown topped by a large beige head that displays impressive persistance and leaves a small to moderate amount of lace in sheets when it eventually departs}
\textcolor{green}{the nose is sweet and spicy and the flavor is malty sweet , accented nicely by honey and by abundant caramel/toffee notes .} there ...... alcohol . 
\textcolor{blue}{the mouthfeel is exemplary ; full and rich , very creamy . mouthfilling with some mouthcoating as well .} drinkability is high ......
\\\hline
Bernoulli & \textcolor{red}{clear , burnished copper-brown topped by a large beige head that displays impressive persistance and} leaves a small to moderate amount of lace in sheets when it eventually departs
the nose is \textcolor{green}{sweet and spicy} and the flavor is malty sweet , accented nicely by honey and by abundant caramel/toffee notes . there ...... alcohol . 
the mouthfeel \textcolor{blue}{is exemplary ; full and rich , very creamy . mouthfilling with} some mouthcoating as well . drinkability is high ......\\\hline
HardKuma & \textcolor{red}{clear , burnished copper-brown topped by a large beige head that displays impressive persistance and leaves a small} to moderate amount of lace in sheets when it eventually departs the nose is \textcolor{green}{sweet and spicy} and the flavor is malty sweet , accented nicely by honey and by abundant caramel/toffee notes . there ...... alcohol . the mouthfeel \textcolor{blue}{is exemplary ; full and rich , very creamy . mouthfilling with} some mouthcoating as well . drinkability is high ...... \\\hline
InfoCal & \textcolor{red}{clear , burnished copper-brown topped by a large beige head that displays impressive persistance} and leaves a small to moderate amount of lace in sheets when it eventually departs \textcolor{green}{the nose is sweet and spicy and the flavor is malty sweet} , accented nicely by honey and by abundant caramel/toffee notes . there ...... alcohol . \textcolor{blue}{the mouthfeel is exemplary ; full and rich , very creamy . mouthfilling with some mouthcoating }as well . drinkability is high ...... \\\hline
InfoCal$-L_\text{adv}$ &\textcolor{red}{clear , burnished copper-brown topped by a large beige head that displays impressive persistance} and leaves a small to moderate amount of lace in sheets when it eventually departs the nose is \textcolor{green}{sweet and spicy} and the flavor is malty sweet , accented nicely by honey and by abundant caramel/toffee notes . there ...... alcohol . \textcolor{blue}{the mouthfeel is exemplary }; full and rich , very creamy . mouthfilling with some mouthcoating as well . drinkability is high ...... \\\hline
InfoCal$-L_\text{lm}$ & \textcolor{red}{clear , burnished} copper-brown topped by a large beige head that displays \textcolor{red}{impressive persistance} and leaves a small to \textcolor{red}{moderate amount of lace} in sheets when it eventually departs the nose is \textcolor{green}{sweet} and \textcolor{green}{spicy} and the flavor is \textcolor{green}{malty sweet} , accented nicely by \textcolor{green}{honey} and by abundant caramel/toffee notes . there ...... alcohol . the mouthfeel is \textcolor{blue}{exemplary} ; \textcolor{blue}{full} and rich , very \textcolor{blue}{creamy} . mouthfilling with some mouthcoating as well . drinkability is high ...... \\
\hline
\end{tabular}
}

\smallskip 
\caption{One example of extracted rationales by different methods. Different colors correspond to different aspects: \textcolor{red}{red}: appearance, \textcolor{green}{green}: smell, and \textcolor{blue}{blue}: palate.}
\label{tab:case}
\end{table*}%

\subsection{Training and Inference}
The total loss function of our model, which takes the generator's role in adversarial training, is shown in Eq.~\ref{eq:G}. The adversarial-related losses are denoted by $L_\text{adv}$. The discriminator is trained by $L_d$ from Eq.~\ref{eq:D}.
\begin{align}
L_\text{adv} &= \lambda_{g}L_g + L_{guide} + \lambda_{mi}L_\text{mi}\\
J_\text{total}&=L_{sp} +\lambda_{ib}L_\text{ib}+L_\text{adv}+\lambda_{lm}L_\text{lm}, \label{eq:G}
\end{align}
where $\lambda_{ib},\lambda_{g},\lambda_{mi}$, and $\lambda_{lm}$ are hyperparameters.

At training time, we optimize the generator loss $J_\text{total}$ and discriminator loss $L_d$ alternately until convergence. The detailed algorithm for training is given in Appendix~\ref{training}  in the extended paper. 
At inference time, we run the selector-predictor model to obtain the prediction and the rationale~$\tilde{\bfz}_\text{sym}$.

\begin{figure}[!t]
\centering
\includegraphics[width=\linewidth]{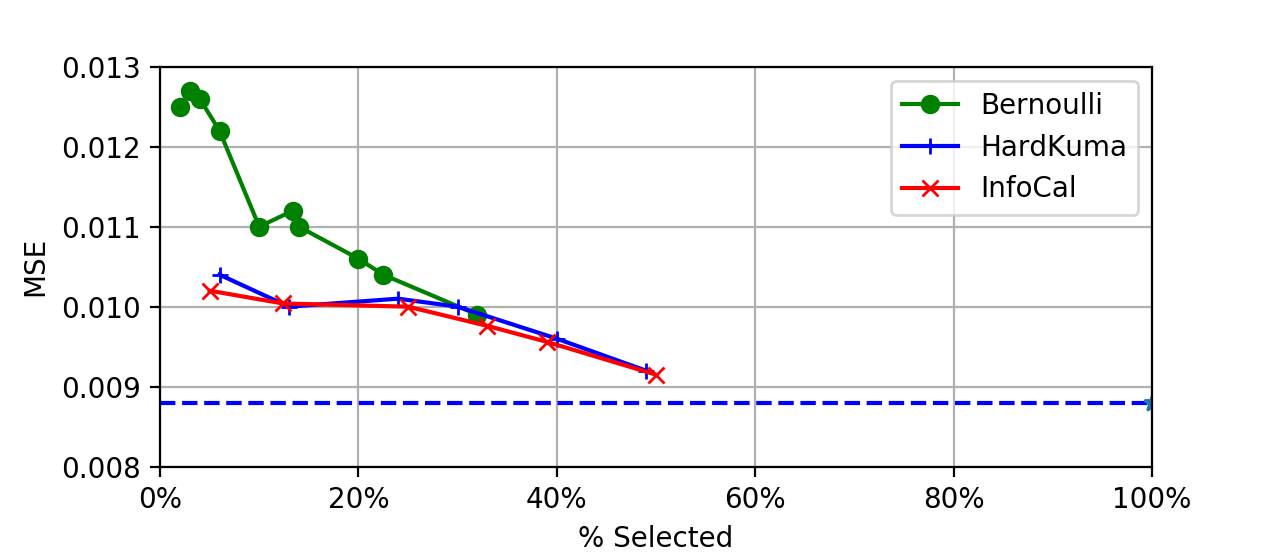}\\

\caption{MSE of all aspects of BeerAdvocate. The blue dashed line represents the full-text baseline (all tokens are selected).}
\label{tab:mse}\vspace*{1ex}
\end{figure}

\section{Experiments}

We performed experiments on two NLP applications: multi-aspect sentiment analysis and legal judgement prediction.
\subsection{Beer Reviews}

\subsubsection{Data.}
To provide a quantitative analysis for the extracted rationales, we use the BeerAdvocate\footnote{\url{https://www.beeradvocate.com/}} dataset~\cite{mcauley2012learning}. This dataset contains instances of human-written multi-aspect reviews on beers. Similarly to \citet{lei2016rationalizing}, we consider the following three aspects: appearance, smell, and palate. \newcite{mcauley2012learning} provide manually annotated rationales for 994 reviews for all aspects, which we use as test set. The detailed data preprocessing and experimental settings are given in Appendix \ref{expsetting}  in the extended paper.


\begin{figure}[!t]
\centering
\includegraphics[width=\linewidth]{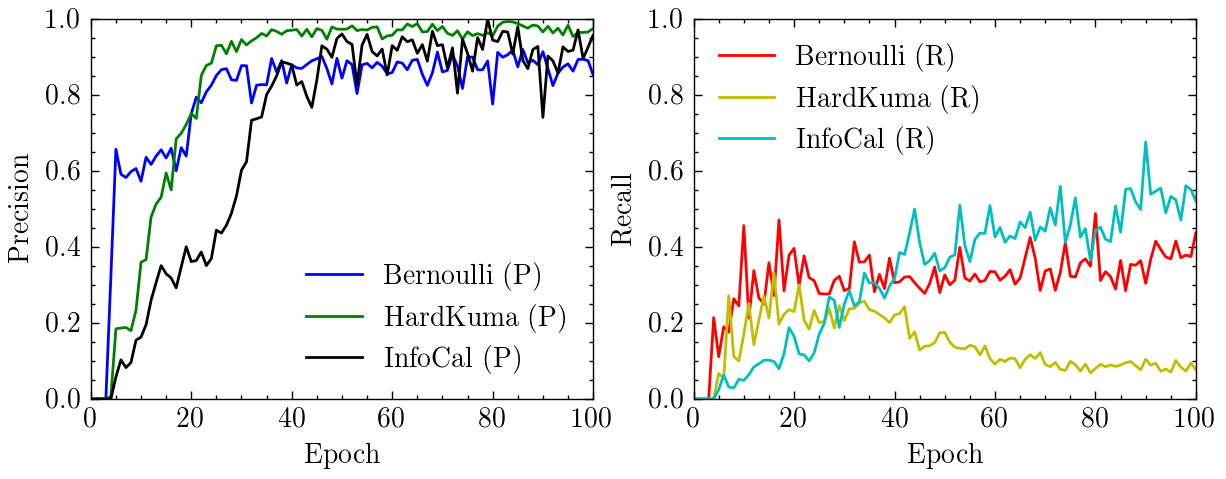}\\

\caption{The precision (left) and recall (right) for rationales  on the smell aspect of the BeerAdvocate test set. } 
\label{tab:pr}\vspace*{1ex}
\end{figure}

\subsubsection{Evaluation Metrics and Baselines.}
For the evaluation of the selected tokens as rationales, we use precision, recall, and F1-score. Typically, precision is defined as the percentage of selected tokens that also belong to the human-annotated rationale. Recall is the percentage of human-annotated rationale tokens that are selected by our model. The predictions made by the selected rationale tokens are evaluated using the mean-square error (MSE).

We compare our method with the following baselines:
\begin{itemize}
\item Attention~\cite{lei2016rationalizing}: This method calculates attention scores over the tokens and selects top-k percent tokens as the rationale.
\item Bernoulli~\cite{lei2016rationalizing}: This method uses a selector network to calculate a Bernoulli distribution for each token, and then samples the tokens from the distributions as the rationale. 
\item HardKuma~\cite{bastings2019interpretable}:  This method replaces the Bernoulli distribution by a Kuma distribution to facilitate differentiability.
\end{itemize}

The details of the choice of neural architecture for each module of our model, as well as the training setup are given in Appendix~\ref{expsetting}  in the extended paper. 


\subsubsection{Results.}
The rationale extraction performances are shown in Table~\ref{tab:beer_rational}. The precision values for the baselines are directly taken from \cite{bastings2019interpretable}. We use their source code for the Bernoulli\footnote{\url{https://github.com/taolei87/rcnn}} and HardKuma\footnote{\url{https://github.com/bastings/interpretable_predictions}} baselines. 
%
We trained these baseline for 50 epochs and selected the models with the best recall on the dev set when the precision was equal or larger than the reported dev precision. For fair comparison, we used the same stopping criteria for InfoCal (for which we fixed a threshold for the precision at 2\% lower than the previous state-of-the-art). 

We also conducted ablation studies: (1) we removed the adversarial loss and report the results in the line  InfoCal$-L_\text{adv}$, and (2)~we removed the LM regularizer and report the results in the line InfoCal$-L_\text{lm}$. 
 
In Table~\ref{tab:beer_rational}, we see that, although Bernoulli and HardKuma achieve very high precisions, their recall scores are significantly low. 
In comparison, our method InfoCal significantly outperforms the previous methods in the recall scores for all the three aspects of the BeerAdvocate dataset (we performed Student's t-test, $p<0.01$). Also, all the three F-scores of InfoCal are a new state-of-the-art performance.

In the ablation studies, we see that when we remove the adversarial information calibrating structure, namely, for InfoCal$-L_\text{adv}$, the recall scores decrease significantly in all the three aspects. This shows that our guider model is critical for the increased performance. 
Moreover, when we remove the LM regularizer, we find a significant drop in both precision and recall, in the line InfoCal$-L_\text{lm}$. This highlights the importance of semantical fluency of rationales, which are encouraged by our LM regularizer.  

We also replace the LM regularizer with the regularizer used in the  HardKuma method with all the other parts of the model unchanged, denoted InfoCal(HK) in Table~\ref{tab:beer_rational}. We found that the recall and F-score of InfoCal outperforms InfoCal(HK), which shows the effectiveness of our LM regularizer.

\begin{table*}[!t]
\begin{center}
\resizebox{0.9\linewidth}{!}{
\begin{tabular}{l l |ccccc| ccccc |ccccc}
\toprule[1.0pt]
\multirow{2}{*}{Small}&Tasks &  \multicolumn{5}{c|}{Law Articles} & \multicolumn{5}{c|}{Charges} & \multicolumn{5}{c}{Terms of Penalty}\\
\cmidrule[0.5pt]{2-17}
 &Metrics& Acc & MP & MR & F1 & \%S & Acc & MP & MR & F1 &\%S & Acc & MP & MR & F1 &\%S   \\
\midrule[0.5pt]
\multirow{8}{*}{Single} &Bernoulli &0.812 & 0.726 & 0.765 & 0.756 &100 & 0.810 & 0.788 & 0.760 & 0.777  & 100 &  0.331 & 0.323 & 0.297 & 0.306 & 100\\
&Bernoulli &0.755 & 0.701 & 0.737 & 0.728 &14 & 0.761 & 0.753 & 0.739 & 0.754  & 14 &  0.323 & 0.308 & 0.265 & 0.278 & 30\\
  &HardKuma & 0.807 & 0.704 & 0.757 & 0.739 &100&0.811 & 0.776 & 0.763 & 0.776   &100 &  0.345 & 0.355 & 0.307 & 0.319& 100\\
  &HardKuma & 0.783 & 0.706 & 0.735 & 0.729 &14&0.778 & 0.757 &0.714 &0.736 &14 &  0.340 & 0.328 &0.296 & 0.309 & 30\\
 \cline{2-17}
 &InfoCal              & \uwave{0.834}  &\uwave{0.744} &\uwave{0.776} &\uwave{0.786} &14 &\uwave{0.849}  &\uwave{0.817} &\uwave{0.798} &\uwave{0.813} &14  &\uwave{0.358} &\uwave{0.372} & \uwave{0.335}&\uwave{0.337} &30\\
 &InfoCal$-L_\text{adv}$ & 0.826  &0.739 &0.774 &0.777 &14 &0.845  &0.804 &0.781 &0.797  &14  &0.351 &0.374 & 0.329&0.330 &30\\
 &InfoCal$-L_\text{adv}\!\!-\!\!L_\text{ib}$ & \textbf{0.841}  & \textbf{0.759}& \textbf{0.785}&\textbf{0.793} &100  &  \textbf{0.850}  & \textbf{0.820}& \textbf{0.801}&\textbf{0.814}  &100 & \textbf{0.368}&\textbf{0.378} &\textbf{0.341}  & \textbf{0.346}&100\\
 &InfoCal$-L_\text{lm}$ & 0.822  &0.723 &0.768 &0.773 &14 &0.843  &0.796 &0.770 &0.772 &14  &0.347 &0.361 & 0.318&0.320 &30\\
 \midrule[0.5pt]
\multirow{3}{*}{Multi} &FLA & 0.803& 0.724& 0.720 &0.714 &$-$&0.767& 0.758& 0.738& 0.732&$-$ &0.371& 0.310 &0.300 &0.299&$-$\\
 &TOPJUDGE & 0.872 & 0.819 &0.808 &0.800 &$-$&0.871 &0.864& 0.851 &0.846&$-$& 0.380 &0.350 &0.353&0.346&$-$\\
 &MPBFN-WCA &\underline{0.883} &\underline{0.832}& \underline{0.824} &\underline{0.822}&$-$ &\underline{0.887} &\underline{0.875} &\underline{0.857} &\underline{0.859}&$-$&\underline{ 0.414} &\underline{0.406} &\underline{0.369}& \underline{0.392}&$-$\\

\midrule[1.0pt]
\multirow{2}{*}{Big}&Tasks &  \multicolumn{5}{c|}{Law Articles} & \multicolumn{5}{c|}{Charges} & \multicolumn{5}{c}{Terms of Penalty}\\
\cmidrule[0.5pt]{2-17}
 &Metrics& Acc & MP & MR & F1 &\%S& Acc & MP & MR & F1&\%S & Acc & MP & MR & F1 &\%S\\
\midrule[0.5pt]
 \multirow{8}{*}{Single} &Bernoulli & 0.876    &0.636   & 0.388 &0.625 &100 & 0.857      &0.643    &0.410   &0.569 &100 & 0.509    &0.511   &0.304  &0.312 & 100\\
 &Bernoulli & 0.857    &0.632   & 0.374 &0.621 &14 & 0.848      &0.635    &0.402   &0.543 &14 & 0.496    &0.505   &0.289  &0.306 & 30\\
  &HardKuma &  0.907 & 0.664 & 0.397 & 0.627 & 100&  0.907 & 0.689 & 0.438 & 0.608&100 &   0.555 & 0.547 & 0.335 & 0.356&100\\
  &HardKuma &  0.876 & 0.645 & 0.384 & 0.609 & 14&  0.892 & 0.676 & 0.425 & 0.587&14 &   0.534 & 0.535 & 0.310 & 0.334&30\\
 \cline{2-17} 
 &InfoCal & \uwave{0.956}&\uwave{0.852}& \uwave{0.742}& \textbf{0.805} & 20 &\uwave{0.955}&\uwave{0.868}& \textbf{0.788}&\textbf{0.820} &20 & 0.556&\textbf{0.519}&0.362&0.372 &30  \\
 &InfoCal$-L_\text{adv}$ & 0.953 & 0.844& 0.711&0.782&20  &0.954& 0.857 & 0.772 &0.806& 20&0.552&0.490& 0.353& 0.356 &30\\
 &InfoCal$-L_\text{adv}\!\!-\!\!L_\text{ib}$ &  \textbf{0.959} & \textbf{0.862} & \textbf{0.751}&0.791&100  &\textbf{0.957}&\textbf{0.878}&0.776&0.807&100 &\textbf{0.584}& \textbf{0.519}& \textbf{0.411}&\textbf{0.427} &30\\
 &InfoCal$-L_\text{lm}$ & 0.953 & 0.851& 0.730 & 0.775& 20 & 0.950& 0.857&0.756& 0.789 &20 &0.563& 0.486& 0.374& 0.367& 30\\
\midrule[0.5pt]
\multirow{3}{*}{Multi}&FLA & 0.942&0.763&0.695&0.746&$-$&0.931&0.798&0.747&0.780&$-$&0.531&0.437&0.331&0.370&$-$\\
&TOPJUDGE&0.963&0.870&0.778&0.802&$-$&0.960&0.906&0.824&0.853&$-$&0.569&0.480&0.398&0.426&$-$\\
&MPBFN-WCA&\underline{0.978}&\underline{0.872}&\underline{0.789}&\underline{0.820}&$-$&\underline{0.977}&\underline{0.914}&\underline{0.836}&\underline{0.867}&$-$&\underline{0.604}&\underline{0.534}&\underline{0.430}&\underline{0.464}&$-$\\
\bottomrule[1.0pt]
\end{tabular}
}
\end{center}
\caption{The overall performance on the CAIL2018 dataset (Small and Big). The results from previous works are directly quoted from \newcite{yang2019legal}, because we share the same experimental settings, and hence we can make direct comparisons. \%S represents the selection percentage (which is determined by the model). ``Single'' represents single-task models, ``Multi'' represents multi-task models. The best performance is in bold. The red numbers mean that they are less than the best performance by no more than $0.01$. The underlined numbers are the state-of-the-art performances, all of which are obtained by multi-task models.}
\label{tab:overall_law}
\end{table*}%

We further show the relation between a model's performance on predicting the final answer and the rationale selection percentage (which is determined by the model) in Fig.~\ref{tab:mse}, as well as the relation between precision/recall and training epochs in Fig.~\ref{tab:pr}. The rationale selection percentage is influenced by $\lambda_\text{ib}$. According to Fig.~\ref{tab:mse}, our method InfoCal achieves a similar prediction performance compared to previous works, and does slightly better than HardKuma for some selection percentages. Fig.~\ref{tab:pr} shows the changes in precision and recall with training epochs. We can see that our model achieves a similar precision after several training epochs, while significantly outperforming the previous methods in recall, which proves the effectiveness of our proposed method.
 
Table~\ref{tab:case} shows an example of rationale extraction. Compared to the rationales extracted by Bernoulli and HardKuma, our method provides more fluent rationales for each aspect. For example, unimportant tokens like ``and'' (after ``persistance'', in the Bernoulli method), and ``with'' (after ``mouthful'', in the HardKuma method) were selected just because they are adjacent to important ones.

\subsection{Legal Judgement Prediction}
\subsubsection{Datasets and Preprocessing.}
We use the CAIL2018 data\-set\footnote{ \url{https://cail.oss-cn-qingdao.aliyuncs.com/CAIL2018_ALL_DATA.zip}}~\cite{zhong-etal-2018-legal} for three tasks on legal judgment prediction.
The dataset consists of criminal cases published by the Supreme People's Court of China.\footnote{\url{http://cail.cipsc.org.cn/index.html}} To be consistent with previous works, we used two versions of CAIL2018, namely, CAIL-small (the exercise stage data) and CAIL-big (the first stage data). 

The instances in CAIL2018 consist of a \textit{fact description} and three kinds of annotations: \textit{applicable law articles}, \textit{charges}, and \textit{the penalty terms}. Therefore, our three tasks on this dataset consist of predicting (1)~law articles, (2)~charges, and (3)~terms of penalty according to the given fact description. 
The detailed experimental settings are given in Appendix~\ref{expsetting}  in the extended paper.

 

\subsubsection{Overall Performance.}
We again compare our method with the Bernoulli~\cite{lei2016rationalizing} and the  HardKuma~\cite{bastings2019interpretable} methods on rationale extraction. These two methods are both single-task models, which means that we train a model separately for each task. 
We also compare our method with three multi-task methods listed as follows:
\begin{itemize}
\item FLA~\cite{luo-etal-2017-learning} uses an attention mechanism to capture the interaction between fact descriptions and applicable law articles.
\item TOPJUDGE~\cite{zhong-etal-2018-legal} uses a topological architecture to link different legal prediction tasks together, including the prediction of law articles, charges, and terms of penalty.
\item MPBFN-WCA~\cite{yang2019legal} uses a backward verification to verify upstream tasks given the results of downstream tasks.
\end{itemize}
The results 
are listed in Table~\ref{tab:overall_law}.  

On CAIL-small, we observe that it is more difficult for the single-task models to outperform multi-task methods. This is likely due to the fact that the tasks are related, and learning them together can help a model to achieve better performance on each task separately. After removing the restriction of the information bottleneck,  InfoCal$-L_\text{adv}\!\!-\!\!L_\text{ib}$ achieves the best performance in all tasks, however,  it selects all the tokens in the review. 
When we restrict the number of selected tokens to $14\%$ (by tuning the hyperparameter $\lambda_\text{ib}$), InfoCal (in red) only slightly drops in all evaluation metrics, and it already outperforms Bernoulli and HardKuma, even if they have used all tokens. This  means that the $14\%$ selected tokens are very important to the predictions. We observe a similar phenomenon for CAIL-big. Specifically, InfoCal outperforms InfoCal$-L_\text{adv}\!\!-\!\!L_\text{ib}$  in some evaluation metrics, such as the F1-score of law article prediction and charge prediction tasks.

\subsubsection{Rationales.}
The CAIL2018 dataset does not contain annotations of rationales. Therefore, we conducted human evaluation for the extracted rationales. Due to limited budget and resources, we sampled 300 examples for each task. We randomly shuffled the rationales for each task and asked six undergraduate students from Peking University to evaluate them. The human evaluation is based on three metrics: usefulness (U), completeness (C), and fluency (F); each scored from $1$ (lowest) to~$5$. The scoring standard for human annotators is given in Appendix \ref{human}  in the extended paper.  

The human evaluation results are  shown in Table~\ref{tab:he}. We can see that our proposed method outperforms previous methods in all metrics. Our inter-rater agreement is acceptable by Krippendorff's rule~(\citeyear{krippendorff2004content}), which is shown in Table~\ref{tab:he}.

\begin{figure}[!t]
\centering
\includegraphics[width=\linewidth]{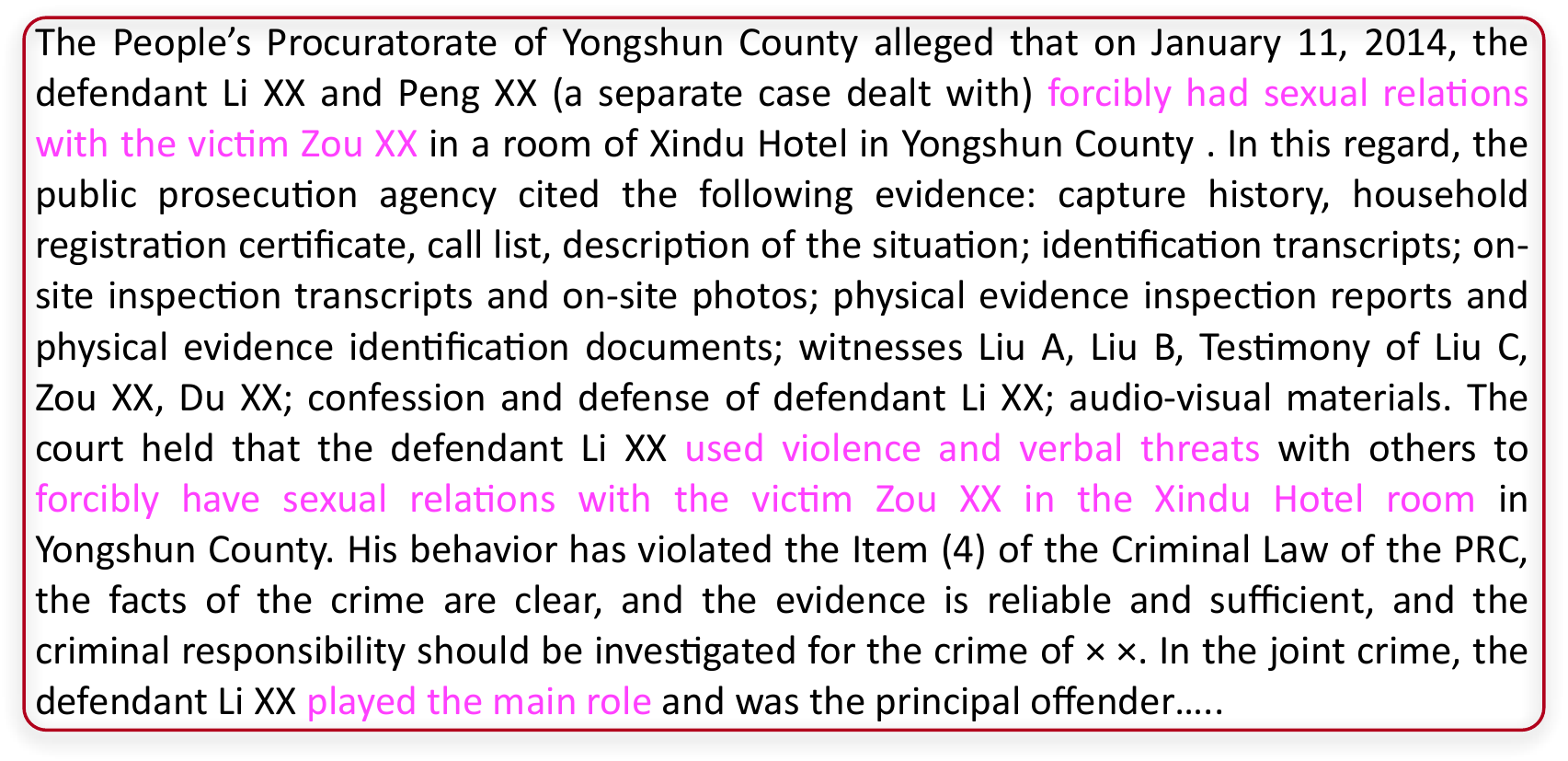}\\
\smallskip 
\caption{An example of extracted rationale for charge prediction. The correct charge is ``Rape". The original fact description is in Chinese, we have translated it to English. It is easy to see that the extracted rationales are very helpful in making the charge prediction.}
\label{fig:lawcase}\vspace*{1ex}
\end{figure}

A sample case of extracted rationales in legal judgement is shown in Fig.~\ref{fig:lawcase}. We observe that our method selects all the useful information for the charge prediction task, and the selected rationales are formed of continuous and fluent sub-phrases.

\begin{table}[!t]
\centering
\resizebox{\linewidth}{!}{
\begin{tabular}{|l|c|c|c|c|c|c|c|c|c|}
\hline
         & \multicolumn{3}{|c|}{Law} &  \multicolumn{3}{|c|}{Charges} & \multicolumn{3}{|c|}{ToP} \\\cline{2-10}
          & U & C &F & U & C & F&U & C  & F \\\hline
Bernoulli &4.71  &2.46 &3.45 & 3.67 &2.35 &3.45 &3.35 &2.76&3.55 \\\hline
HardKuma  &4.65  &3.21 &3.78 & 4.01 &3.26&3.44  &3.84 &2.97&3.76\\\hline
InfoCal  &\textbf{4.72} &\textbf{3.78} &\textbf{4.02} &\textbf{4.65} & \textbf{3.89}&\textbf{4.23} &\textbf{4.21} &\textbf{3.43}&\textbf{3.97}\\\hline\hline
 $\alpha$&0.81 & 0.79&0.83&0.92&0.85&0.87&0.82&0.83&0.94\\\hline
\end{tabular}
}
\smallskip 
\caption{Human evaluation on the CAIL2018 dataset. ``ToP" is the abbreviation of ``Terms of Penalty". The metrics are: usefulness (U), completeness (C), and fluency (F), each scored from $1$ to~$5$. Best performance is in bold. $\alpha$ represents Krippendorff's alpha values. }
\label{tab:he}\vspace*{1ex}
\end{table}%

\section{Related Work}
Explainability is currently a key bottleneck of deep-lear\-ning-based approaches. 
The model proposed in this work belongs to the class of self-explanatory models, which contain an explainable structure in the model architecture, thus providing explanations for their predictions. Self-explanatory models can use different types of explanations, such as feature-based explanations \cite{lei2016rationalizing, yoon2018invase,chen2018learning,yu-etal-2019-rethinking,carton-etal-2018-extractive} and natural language explanations 
\cite{DBLP:conf/eccv/HendricksARDSD16, esnli, zeynep, cars}. Our model uses feature-based explanations.  

Self-explanatory models with feature-based explanations can be further divided into two branches. The first branch is formed of representation-inter\-pretable approaches, which map specific features into latent spaces and then use the latent variables to control~the outcomes of the model, such as disentangling methods~\cite{chen2016infogan,sha2021multi}, information bottleneck methods~\cite{tishby2000information}, and constrained generation~\cite{sha-2020-gradient}. The second branch consists of architecture-in\-ter\-pretable models, such as attention-based models \cite{zhang2018top,sha2016reading,sha2018order,sha2018multi,liu2017table}, neural Turing machines~\cite{collier2018im,xia-etal-2017-progressive,sha2020estimating}, capsule networks~\cite{sabour2017dynamic}, and energy-based models~\cite{grathwohl2019your}. Among them, attention-based models have an important extension, that of sparse feature learning, which implies learning to extract a subset of features that are most informative for each example. Most of the sparse feature learning methods use a selector-predictor architecture. Among them, L2X \cite{chen2018learning} and INVASE~\cite{yoon2018invase} make use of information theories for feature selection, while CAR~\cite{chang2019game} extracts useful features in a game-theoretic approach.

In addition, rationale extraction for NLP usually raises one desideratum for the extracted subset of tokens: rationales need to be fluent subphrases instead of separate tokens. To this end, \newcite{lei2016rationalizing} proposed a non-differentiable regularizer to encourage selected tokens to be consecutive, which can be optimized by REINFORCE-style methods~\cite{williams1992simple}. \newcite{bastings2019interpretable} proposed a differentiable regularizer using the Hard Kumaraswamy distribution; however, this still does not consider the difference in the importance of different adjacent token pairs.




Our adversarial calibration method is inspired by distilling methods~\cite{hinton2015distilling}. 
Distilling methods are usually applied to compress large models into small models while keeping a comparable performance. For example, TinyBERT~\cite{jiao2019tinybert} is a distillation of BERT~\cite{devlin-etal-2019-bert}. Our method is different from distilling methods, because we calibrate the final feature vector instead of the softmax prediction. 

The information bottleneck~(IB) theory is an important basic theory of neural networks ~\cite{tishby2000information}. It originated in information theory and has been widely used as a theoretical framework in analyzing deep neural networks~\cite{tishby2015deep}. For example, \newcite{li-eisner-2019-specializing} used IB to compress word embeddings in order to make them contain only specialized information, which leads to a much better performance in parsing tasks.

Adversarial methods, which had been widely applied in image generation~\cite{chen2016infogan} and text generation~\cite{yu2017seqgan}, usually have a discriminator and a generator. The discriminator receives pairs of instances from the real distribution and from the distribution generated by the generator, and it is trained to differentiate between the two. The generator is trained to fool the discriminator~\cite{goodfellow2014generative}. Our information calibration method generates a dense feature vector using selected symbolic features, and the discriminator is used for measuring the calibration extent. 

\section{Summary and Outlook}

In this work, we proposed a novel method to extract rationales for neural predictions. Our method uses an adversarial-based technique to make a selector-predictor model learn from a guider model. In addition, we proposed a novel regularizer based on language models, which makes the extracted rationales semantically fluent. 
The experimental results showed that our method improves the selection of rationales by a large margin. 

As future work, the main architecture of our model can be directly applied to other domains, e.g., images or tabular data. However, it remains an open question what would be a good regularizer for these domains.  

\section*{Acknowledgments}
 This work was supported by the EPSRC grant ``Unlocking the Potential of AI for English Law'', a JP Morgan PhD Fellowship, the Alan Turing Institute under the EPSRC grant EP/N510129/1, and the AXA Research Fund. We also acknowledge the use of Oxford's Advanced Research Computing (ARC) facility, of the EPSRC-funded Tier 2 facility
JADE (EP/P020275/1), and of GPU computing support by Scan Computers International Ltd.

\bibliography{refs}
\clearpage
\appendix
\section*{Appendices}
\section{Proofs}
\subsection{Derivation of $I(\tilde{\bfz}_\text{sym},y)$}\label{the:0}
\begin{theorem}
Minimizing $-I(\tilde{\bfz}_\text{sym},y)$ is 
equivalent to minimizing $L_{sp}$.
\end{theorem}
\begin{proof}
\begin{equation}\label{eq:Izy}
\begin{aligned}
I(\tilde{\bfz}_\text{sym},y)&=\mathbb E_{\tilde{\bfz}_\text{sym},y}\Big[\frac{p(y|\tilde{\bfz}_\text{sym})}{p(y)}\Big]\\
&=\mathbb E_{\tilde{\bfz}_\text{sym},y}p(y|\tilde{\bfz}_\text{sym}) - \mathbb E_{\tilde{\bfz}_\text{sym},y}p(y).\\[0.5ex] 
\end{aligned}
\end{equation}
We omit $\mathbb E_{\tilde{\bfz}_\text{sym},y}p(y)$, because it is a constant, therefore, minimizing  Eq.~\ref{eq:Izy} is equivalent to minimizing the following term:
\begin{equation}\label{eq:Ezy}
\mathbb E_{\tilde{\bfz}_\text{sym},y}p(y|\tilde{\bfz}_\text{sym}).
\end{equation}
As the training pair $(\bfx, y)$ is sampled from the training data, and $\tilde{\bfz}_\text{sym}$ is sampled from $\text{Sel}(\tilde{\bfz}_\text{sym}|\bfx)$, we have that 
\begin{equation}\label{eq:Ezy1}
\begin{aligned}
\mathbb E_{\tilde{\bfz}_\text{sym},y}p(y|\tilde{\bfz}_\text{sym}) &= 
\mathbb E_{\bfx,y}p(y|\tilde{\bfz}_\text{sym})p(\tilde{\bfz}_\text{sym}|\bfx)\\
&=E_{\bfx,y}p(y|\bfx).\\
\end{aligned}
\end{equation}
We can give each $p(y|\bfx)$ in Eq.~\ref{eq:Ezy1} a $-\log$ to arrive to $-I(\tilde{\bfz}_\text{sym},y)$. Then, it is not difficult to see that 
$-I(\tilde{\bfz}_\text{sym},y)$ has exactly the same form as $L_\text{sp}$.
\end{proof}

\subsection{Derivation of Equation~\ref{eq:mi}} \label{proof:mi}
\begin{align}
L_\text{mi}&=I(\bfx, \bfz_\text{nero})\\
 &= \mathbb E_{\bfx, \bfz_\text{nero}}\Big[\log\frac{p(\bfz_\text{nero}|\bfx)}{p(\bfz_\text{nero})}\Big]\\
 &= \mathbb E_{\bfz_\text{nero}}p(\bfx)\Big[\log\frac{p(\bfz_\text{nero}|\bfx)}{p(\bfz_\text{nero})}\Big]\\
&\le \mathbb E_{\bfz_\text{nero}}\Big[\log\frac{p(\bfz_\text{nero}|\bfx)}{p(\bfz_\text{nero})}\Big]\\
&=0.5(\mu^2+\sigma^2-1-2\log\sigma).
\end{align}

\subsection{Proof of Theorem~\ref{the:1}}\label{proof:1}
\setcounter{theorem}{0}
 \begin{theorem}
If the following is satisfied for all $i,j$: 
\begin{itemize}
    \item $m'_i<\epsilon \ll 1-\epsilon< m_i$, ($0<\epsilon<1$), and 
    \item $\big|p(m'_ix_i|x_{<i})-p(m'_jx_j|x_{<j})\big|<\epsilon$,
\end{itemize}
then the following two inequalities hold:\\
 (1) $L_\text{lm}(\ldots,m_k,\ldots, m'_{n})<L_\text{lm}(\ldots,m'_k,\ldots, m_{n})$.\\
 (2) $L_\text{lm}(m_1, \ldots,m'_k,\ldots)>L_\text{lm}(m'_1,\ldots,m_k,\ldots)$.
\end{theorem}
\begin{proof}
By Eq.~\ref{eq:lm}, we have:\\[0.5ex] 
\begin{equation}
\begin{small}
\begin{aligned}
    &L_\text{lm}(\ldots,m'_k,\ldots, m_{n})=-\Big[\sum_{i\neq k,k+1}m_{i-1}\log P(m_ix_i|x_{<i}) \\
    &+ m_{k-1}\log P(m'_kx_k|x_{<k})+m'_k\log P(m_{k+1}x_{k+1}|x_{<k+1})\Big]. 
\end{aligned}
\end{small}
\end{equation}
Therefore, we have the following equation:\\[0.5ex] 
\begin{equation}\label{eq:pr:1}
\begin{aligned}
    &L_\text{lm}(\ldots,m'_k,\ldots, m_{n})-L_\text{lm}(\ldots,m_k,\ldots, m'_{n})\\
    =&-m_{k-1}\log p(m'_k)+m_{k-1}\log p(m_k)\\
     &-m'_k\log p(m_{k+1})+m_k\log p(m_{k+1})\\
     &-m_{n-1}\log p(m_n)+m_{n-1}\log p(m'_n)\\
     &- m_n\log p(m'_{k+1}) + m'_n\log p(m'_{n+1}),\\[0.5ex] 
\end{aligned}
\end{equation}
where, for simplicity, we use the abbreviation $p(m_k)$ to represent $p(m_kx_k|x_{<k})$.

We also have that \\[0.5ex] 
\begin{align}
&- m_n\log p(m'_{k+1})+m_{n-1}\log p(m'_n)\\
&=(m_{n-1}-m_{n})\log p(m'_{k+1}) - m_{n-1}\log\frac{p(m'_{k+1})}{p(m'_n)}\\
&\ge \epsilon\log p(m'_{k+1})-\epsilon
\end{align}
Since ${p(m_k)}_k$ are expected to have large probability values in the language model training process, we have that $p(m_k)>\delta$, and, therefore, $-|\log\delta|<\log\frac{p(m_{k+1})}{p(m_n)}<|\log\delta|$.

Hence, we have that:\\[0.5ex] 
\begin{align}
&- m_{n-1}\log p(m_n)+m_k\log p(m_{k+1})\\
&=(m_k-m_{n-1})\log p(m_{k+1}) + m_{n-1}\log\frac{p(m_{k+1})}{p(m_n)}\\
&\ge \epsilon\log p(m_{k+1})-|\log\delta|\ge(\epsilon-1)|\log\delta|.\\[0.5ex] 
\end{align}
Similarly, $-m'_k\log p(m_{k+1})+m_{k-1}\log p(m_k)\ge (1-2\epsilon)\log p(m_k)+m'_k\log\frac{p(m_{k})}{p(m_{k+1})}\ge (1-3\epsilon) |\log\delta|$.\\[0.5ex]  

Therefore, the lower bound of the expression in Eq.~\ref{eq:pr:1} is:\\[0.5ex] 
\begin{equation}
\begin{aligned}
\inf&=-(1-\epsilon)\log p(m'_k)+ \epsilon\log p(m'_{n+1})\\
    &\quad+\epsilon\log p(m'_{k+1})-2\epsilon|\log\delta|-\epsilon\\
	&\ge -(1-3\epsilon)\log p(m'_k)-4\epsilon|\log\delta|-\epsilon >0.\\[0.5ex] 
\end{aligned}
\end{equation}
This proves the statement of the theorem.
\end{proof}

\begin{table*}[!h]
\centering
\resizebox{\linewidth}{!}{
\begin{tabular}{|p{10cm}|p{10cm}|}
\hline
\multicolumn{1}{|c|}{Gold} & \multicolumn{1}{c|}{InfoCal}\\
\hline
\textcolor{red}{dark black with nearly no light at all shining through on this one . rich tan colored head of about two inches quickly settled down to about a half inch of tan that thoroughly coated the inside of the glass .} this was what the style is all about \textcolor{green}{the aroma was just loaded down with coffee . rich  notes of mocha mixes in with a rich , and sweet coffee note . a tiny bit of bitterness and an earthy flare lying down underneath of it , but the majority of this one was hands down , rich brewed coffee .} the flavor was more of the same . \textcolor{blue}{rich notes just rolled over the tongue in waves and thoroughly coated the inside of the mouth .} sweet with touches of chocolate and vanilla to highlight the coffee notes &
\textcolor{red}{dark black with nearly no light at all shining through on this one . rich tan colored head of about two inches quickly settled down} to about a half inch of tan that thoroughly coated the inside of the glass . this was what the style is all about \textcolor{green}{the aroma was just loaded down with coffee . rich  notes of mocha mixes in with a rich , and sweet coffee note .} a tiny bit of \textcolor{green}{bitterness and an earthy flare lying down underneath of it }, but the majority of this one was hands down , rich brewed coffee . the flavor was more of the same . \textcolor{blue}{rich notes just rolled over the tongue in waves and thoroughly coated the inside of the mouth} . sweet with touches of chocolate and vanilla to highlight the coffee notes\\ 
\hline
\textcolor{red}{clear copper colored brew , medium cream colored  head . }\textcolor{green}{floral hop nose , caramel malt .} caramel malt front dominated by a nice floral hop backround . grapefruit tones . very tasty hops run the show with this brew .\textcolor{blue}{ thin to medium mouth }. not a bad choice if you 're looking for a nice hop treat .
&
\textcolor{red}{clear copper colored brew , medium cream colored}  head  \textcolor{green}{. floral hop nose , caramel malt .} caramel malt front dominated by a nice floral hop backround . grapefruit tones . very tasty hops run the show with this brew .\textcolor{blue}{ thin to medium mouth }. not a bad choice if you 're looking for a nice hop treat .
\\
\hline
12oz bottle into my pint glass . \textcolor{red}{looks decent , a brown color ( imagine that ! ) with a tan head . nothing bad , nothing extraordinary . }\textcolor{green}{smell is nice , slight roast , some nuttiness , and hint of hops . pretty much to-style . }taste is good but a little underwhelming . toffee malt , some slight roast gives chocolate impressions . hoppiness is mild and earthy . just a touch of bitterness . pretty nondescript overall , but nothing offensive . \textcolor{blue}{mouthfeel is good , medium body and light carb give a creamy finish . }drinkability was nice . i would try this again but wo n't be seeking it out .
&
12oz bottle into my \textcolor{red}{pint} glass . \textcolor{red}{looks decent , a brown color ( imagine that ! ) with a tan head .} nothing bad , nothing extraordinary . \textcolor{green}{smell is nice , slight roast , some nuttiness ,} and hint of hops . pretty much to-style . taste is good but a little underwhelming . toffee malt , some slight roast gives chocolate impressions . hoppiness is mild and earthy . just a touch of bitterness . pretty nondescript overall , but nothing offensive . \textcolor{blue}{mouthfeel is good , medium body and light carb give a creamy} finish . drinkability was nice . i would try this again but wo n't be seeking it out .\\\hline
\end{tabular}
}
\smallskip 
\caption{More instances from the BeerAdvocate dataset. In red the rationales for the appearance aspect, in green the rationales for the smell aspect, and in blue the rationales for the palate aspect. }
\label{tab:beermorecase}
\end{table*}%

\begin{figure*}[!h]
\begin{center}
\includegraphics[width=\linewidth]{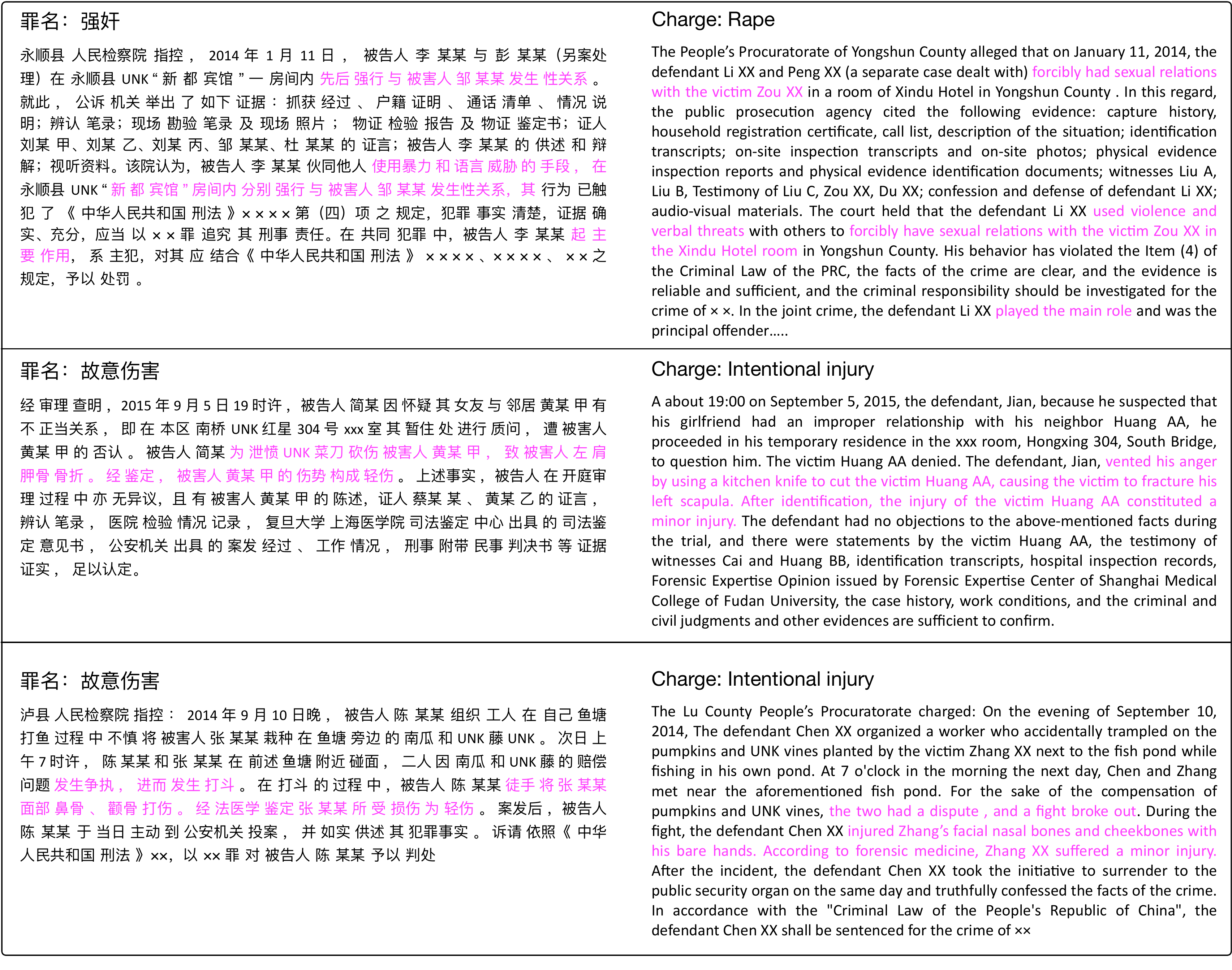}\\
\caption{More instances from the CAIL2018 dataset. Left: the fact description (in Chinese). Right: the corresponding English translation of the fact description. In pink is the selected rationales.}
\label{tab:lawmorecase}
\end{center}
\end{figure*}

\section{Experimental Settings}\label{expsetting}

\subsection{Beer Reviews}

\subsubsection{The BeerAdvocate dataset.}
The training set of BeerAdvocate contains 220,000 beer reviews, with human ratings for each aspect. 
Each rating is on a scale of $0$ to $5$ stars, and it can be fractional (e.g., 4.5 stars), \newcite{lei2016rationalizing} have normalized the scores to $[0,1]$, and picked ``less correlated'' examples to make a de-correlated subset.\footnote{\url{http://people.csail.mit.edu/taolei/beer/}} For each aspect, there are  80k--90k reviews for training and 10k reviews for validation.

\subsubsection{Model details.}
Because our task is a regression, we make some modifications to our model. First, we replace the \textit{softmax} in Eq.~\ref{eq:pyn} by the \textit{sigmoid} function, and replace the cross-entropy loss in Eq.~\ref{eq:lsp} by a mean-squared error (MSE) loss. Second, for a fair comparison, similar to \newcite{lei2016rationalizing} and \newcite{bastings2019interpretable}, we set all the architectures of selector, predictor, and guider as bidirectional Recurrent Convolution Neural Network (RCNN)~\newcite{lei2016rationalizing}, which performs similarly to an LSTM~\cite{hochreiter1997long} but with $50\%$ fewer parameters.

We search the hyperparameters in the following scopes: $\lambda_\text{ib}\in(0.000, 0.001]$ with step $0.0001$, $\lambda_g\in[0.2,2.0]$ with step $0.2$, $\lambda_\text{mi}\in[0.0, 1.0]$ with step $0.1$, and $\lambda_\text{lm}\in[0.000,0.010]$ with step $0.001$.

The best hyperparameters were found as follows: $\lambda_\text{ib}=0.0003$, $\lambda_g=1$, $\lambda_\text{mi}=0.1$, and $\lambda_\text{lm}=0.005$. 

We set $r_\phi(z_i)$  to $r_\phi(z_i=0)=0.999$ and $r_\phi(z_i=1)=0.001$.

\subsection{Legal Judgement Prediction}
The statistics of CAIL2018 dataset are shown in Table~\ref{tab:cail}.
\begin{table}[!t]
\begin{center}
\begin{tabular}{lcc}
\toprule[1.0pt]
 & CAIL-small & CAIL-big\\
\midrule[0.5pt]
Cases & 113,536 & 1,594,291\\
Law Articles & 105 &183 \\
Charges & 122 & 202\\
Term of Penalty &11 & 11\\
\bottomrule[1.0pt]
\end{tabular}
\end{center}
\caption{Statistics of the CAIL2018 dataset.}
\label{tab:cail}\vspace*{2ex}
\end{table}%

In the dataset, there are also many cases with multiple applicable law articles and multiple charges. To be consistent with previous works on legal judgement prediction \cite{zhong-etal-2018-legal,yang2019legal}, we filter out these multi-label examples. 

We also filter out instances where the charges and law articles occurred less than $100$ times in the dataset (e.g., insulting the national flag and national emblem).
For the term of penalty, we divide the terms into $11$ non-overlapping intervals. These preprocessing steps are the same as in \newcite{zhong-etal-2018-legal} and \newcite{yang2019legal}, making it fair to compare our model with previous models.
 
We use Jieba\footnote{\url{https://github.com/fxsjy/jieba}} for token segmentation, because this dataset is in Chinese. The word embedding size is set to $100$ and is randomly initiated before training. The maximum sequence length is set to $1000$. The architectures of the selector, predictor, and guider are all bidirectional LSTMs. The LSTM's hidden size is set to $100$. $r_\phi(z_i)$ is the sampling rate for each token (0 for selected), which we set to $r_\phi(z_i=0)=0.9$. 

We search the hyperparameters in the following scopes: $\lambda_{ib}\in[0.00, 0.10]$ with step $0.01$, $lambda_{g}\in[0.2,2.0]$ with step $0.2$,  $\lambda_{mi}\in[0.0, 1.0]$ with step $0.1$, $\lambda_{lm}\in[0.000,0.010]$ with step $0.001$.
The best hyperparameters were found to be: $\lambda_{ib}=0.05, \lambda_{g}=1, \lambda_{mi}=0.5, \lambda_{lm}=0.005$ for all the three tasks. 
 
\section{Language Model as Rationale Regularizer} \label{lmvec}
 
Note that in Eq.~\ref{eq:lm}, the target sequence of the language model $P(m_ix_i|x_{<i})$ is formed of vectors instead of symbolic tokens, as in usually the case for language models. Therefore, we make some small changes in the pre-training of the language model. In typical language models, the last layer is:
\begin{equation}\label{eq:lm_lastlayer}
p(x_i|x_{<i})=\frac{\exp{(h_i^\top e_i)}}{\sum_{j\in\mathcal V}\exp{(h_i^\top e_j)}},
\end{equation}
where $h_i$ is the hidden vector corresponding to $x_i$, $e_i$ represents the output vector of $x_j$, and $\mathcal V$ is the vocabulary. When we are modeling the language model in vector form, we only use a bilinear layer to directly calculate the probability in Eq.~\ref{eq:lm_lastlayer}: 
 \begin{equation}
 p(x_i|x_{<i})=\sigma(h_i^\top Me_i),
 \end{equation}
  where $\sigma$ stands for \texttt{sigmoid}, and $M$ is a trainable parameter matrix. To achieve this, we use negative sampling~\cite{mikolov2013distributed} in the training procedure. The language model is pretrained using the following loss:
\begin{equation}\label{eq:pre}
 L_\text{pre}=-\sum_i\Big[\log\sigma(h_i^\top Me_i) - \mathbb E_{j\sim p(x_j)}\log\sigma(h_i^\top Me_j) \Big], 
\end{equation}
where $p(x_j)$ is the occurring probability (in the training dataset) of token $x_j$. 

\section{Training Procedure for InfoCal}\label{training}
The whole training process is illustrated in Algorithm~\ref{algo:1}.

\begin{algorithm}
\SetAlgoLined
 Random initialization\;
 Pre-train language model by Eq.~\ref{eq:pre}\;
 \For{each iteration $i=1,2,\ldots$ }{
 \For{each batch  }{
  Calculate the loss $J_\text{total}$ for the sampler-predictor model and the guider model by Eq.~\ref{eq:G}\;
  Calculate the loss $L_D$ for the discriminator by Eq.~\ref{eq:D}\;
  Update the parameters of selector-predictor model and the guider model\;
  Update the parameters of the discriminator\;
  }
 }
 \caption{Training process}
\label{algo:1}
\end{algorithm}
 
 \section{Human Evaluation Setup}\label{human}
Our annotators were asked the following questions, in order to assess the usefulness, completeness, and fluency of the rationales provided by our model.

\subsection{Usefulness of Rationales}

Q: Do you think the selected tokens/rationale are \textbf{useful} to explain the ground-truth label?

Please choose a score according to the following description. Note that the score is not necessary an integer, you can give intermediate scores, such as $3.2$ or $4.9$ if you deem appropriate.
\begin{itemize}
\item 5: Exactly. I can give the correct label only by seeing the given tokens.
\item 4: Highly useful. Although most of the selected tokens lead to the correct label, there are still several tokens that have no relation to the correct label.
\item 3: Half of them are useful. About half of the tokens can give some hint for the correct label, the rest are nonsense to the label.
\item 2: Almost useless. Almost all of the tokens are useless, but there are still several tokens that are useful. 
\item 1: No Use. I feel very confused about the selected tokens, I don't know which law article/charge/term of penalty the article belongs to.
\end{itemize}

\subsection{Completeness of Rationales}

Q: Do you think the selected tokens/rationale are \textbf{enough} to explain the ground-truth label?

Please choose a score according to the following description. Note that the score is not necessary an integer, you can give intermediate scores, such as $3.2$ or $4.9$, if you deem appropriate.
\begin{itemize}
\item 5: Exactly. I can give the correct label only by the given tokens.
\item 4: Highly complete. There are still several tokens in the fact description that have a relation to the correct label, but they are not selected.
\item 3: Half complete. There are still important tokens in the fact description, and they are in nearly the same number as the selected tokens.
\item 2: Somewhat complete. The selected tokens are not enough. There are still many important tokens in the fact description not being selected. 
\item 1: Nonsense. All of the selected tokens are useless. None of the important tokens is selected.

\end{itemize}
\subsection{Fluency}

Q: How fluent do you think the selected rationale is?  For example: \textit{``He stole an iPhone in the room''} is very fluent, which should have a high score. \textit{``stole iPhone room''} is just separated tokens, which should have a low fluency score.

Please choose a score according to the following description. Note that the score is not necessary an integer, you can give scores like $3.2$ or $4.9$ , if you deem appropriate.
\begin{itemize}
\item 5: Very fluent. 
\item 4: Highly fluent. 
\item 3: Partial fluent. 
\item 2: Very unfluent. 
\item 1: Nonsense.
\end{itemize}

\section{More Examples of Rationales}\label{morecase}

\subsection{BeerAdvocate}
We list more examples of rationales extracted by our model for the BeerAdvocate dataset in Table~\ref{tab:beermorecase}.
\subsection{Legal Judgement Prediction}
More examples of  rationales extracted by our model for the legal judgement tasks are shown in Table~\ref{tab:lawmorecase}. 

\end{document}